\documentclass[11pt,a4paper]{article}
\usepackage[table]{xcolor}
\usepackage{pgfplotstable}

\pgfplotstableset{
    color cells/.style={
        col sep=comma,
        string type,
        postproc cell content/.code={%
                \pgfkeysalso{@cell content=\rule{0cm}{2.4ex}\cellcolor{black!##1}\pgfmathtruncatemacro\number{##1}\ifnum\number>60\color{white}\fi##1}%
                },
        columns/x/.style={
            column name={},
            postproc cell content/.code={}
        }
    }
}

\makeatletter
\def\hlinew#1{%
  \noalign{\ifnum0=`}\fi\hrule \@height #1 \futurelet
   \reserved@a\@xhline}
\makeatother

\usepackage[hyperref]{acl2019}
\usepackage{times}
\usepackage{latexsym}

\usepackage{url}

\usepackage{graphicx}  
\usepackage{CJK}
\usepackage{array}
\usepackage{multirow}
\usepackage{enumitem}
\usepackage{amsmath}
\usepackage{mathrsfs}
\usepackage{amsfonts}

\newcommand{\PreserveBackslash}[1]{\let\temp=\\#1\let\\=\temp}
\newcolumntype{C}[1]{>{\PreserveBackslash\centering}p{#1}}
\newcolumntype{R}[1]{>{\PreserveBackslash\raggedleft}p{#1}}
\newcolumntype{L}[1]{>{\PreserveBackslash\raggedright}p{#1}}

\aclfinalcopy 
\def\aclpaperid{857}

\title{Cost-sensitive Regularization for Label Confusion-aware Event Detection}

\author{Hongyu Lin${}^{1,3}$, Yaojie Lu${}^{1,3}$, Xianpei Han${}^{1,2,*}$, Le Sun${}^{1,2}$ \\
${}^{1}$Chinese Information Processing Laboratory ~ ${}^{2}$State Key Laboratory of Computer Science \\
Institute of Software, Chinese Academy of Sciences, Beijing, China\\
${}^{3}$University of Chinese Academy of Sciences, Beijing, China \\
 {\tt \{hongyu2016,yaojie2017,xianpei,sunle\}@iscas.ac.cn}
}

\date{}

\begin{document}
\maketitle
\begin{abstract}

In supervised event detection, most of the mislabeling occurs between a small number of confusing type pairs, including trigger-NIL pairs and sibling sub-types of the same coarse type. To address this label confusion problem, this paper proposes cost-sensitive regularization, which can force the training procedure to concentrate more on optimizing confusing type pairs. Specifically, we introduce a cost-weighted term into the training loss, which penalizes more on mislabeling between confusing label pairs. Furthermore, we also propose two estimators which can effectively measure such label confusion based on instance-level or population-level statistics. Experiments on TAC-KBP 2017 datasets demonstrate that the proposed method can significantly improve the performances of different models in both English and Chinese event detection.

\end{abstract}

\section{Introduction}
{
\renewcommand{\thefootnote}{\fnsymbol{footnote}}
\footnotetext[1]{Corresponding author.}
}

Automatic event extraction is a fundamental task in information extraction. Event detection, aiming to identify trigger words of specific types of events, is a vital step of event extraction. For example, from sentence ``Mary was injured, and then she died'', an event detection system is required to detect a \textit{Life:Injure} event triggered by ``injured'' and a \textit{Life:Die} event triggered by ``died''.

Recently, neural network-based supervised models have achieved promising progress in event detection~\cite{nguyen2015event,chen2015event,ghaeini2016}. Commonly, these methods regard event detection as a word-wise classification task with one \textit{NIL} class for tokens do not trigger any event. Specifically, a neural network automatically extracts high-level features and then feed them into a classifier to categorize words into their corresponding event sub-types (or \textit{NIL}). Optimization criteria of such models often involves in minimizing cross-entropy loss, which equals to maximize the likelihood of making correct predictions on the training data.

\begin{table}
\centering
\setlength{\abovecaptionskip}{0.2cm}
\setlength{\belowcaptionskip}{-0.5cm}
\pgfplotstabletypeset[color cells]{
x,BC,CT,CR,MT,NIL,CC
BC,41.3,14.4,2.1,1.6,39.0,1.7
CT,8.5,42.7,4.7,2.6,40.6,0.9
CR,5.7,7.3,50.0,1.1,32.3,2.9
MT,3.0,7.7,6.1,28.7,51.3,3.2
}
\caption{Prediction percentage heatmap of triggers with \textit{Contact} coarse type. Row labels are the golden label and the column labels indicate the prediction. BC: Broadcast; CT: Conctact(sub-type); CR: Correspondence; MT: Meet; CC: Other cross coarse-type errors.}
\label{tb:heatmap}
\end{table}

However, we find that in supervised event detection, most of the mislabeling occurs between a small number of confusing type pairs. We refer to this phenomenon as \emph{label confusion}. Specifically, there are mainly two types of label confusion in event detection:
1) trigger/NIL confusion; 2) sibling sub-types confusion.
For example, both \textit{Transaction:Transfer-money} and \textit{Transaction:Transfer-ownership} events are frequently triggered by word ``give''. Besides, in many cases ``give'' does not serve as a trigger word. Table~\ref{tb:heatmap} shows the classification results of a state-of-the-art event detection model~\cite{chen2015event} on all event triggers with coarse type of \textit{Contact} on TAC-KBP 2017 English Event Detection dataset. We can see that the model severely suffers from two types of label confusion mentioned above: more than 50\% mislabeling happens between trigger/NIL decision due to the ambiguity of natural language. Furthermore, the majority of remaining errors are between sibling sub-types of the same coarse type because of their semantic relatedness~\cite{liu2017improving}. Similar results are also observed in other event detection datasets such as ACE2005~\cite{LiuEtal:2018:AAAI2018}. Therefore, it is critical to enhance the supervised event detection models by taking such label confusion problem into consideration.

In this paper, inspired by cost-sensitive learning~\cite{ling2011cost}, we introduce cost-sensitive regularization to model and exploit the label confusion during model optimization, which can make the training procedure more sensitive to confusing type pairs.
Specifically, the proposed regularizer reshapes the loss function of model training by penalizing the likelihood of making wrong predictions with a cost-weighted term. If instances of class $i$ are more frequently misclassified into class $j$, we assign a higher cost to this type pair to make the model intensively learn to distinguish between them. Consequently, the training procedure of models not only considers the probability of making correct prediction, but also tries to separate confusing type pairs with a larger margin. Furthermore, in order to estimate such cost automatically, this paper proposes two estimators based on population-level or instance-level statistics.

We conducted experiments on TAC-KBP 2017 Event Nugget Detection datasets. Experiments show that our method can significantly reduce the errors between confusing type pairs, and therefore leads to better performance of different models in both English and Chinese event detection. To the best of our knowledge, this is the first work which tackles with the label confusion problem of event detection and tries to address it in a cost-sensitive regularization paradigm.

\section{Cost-sensitive Regularization for Neural Event Detection}
\subsection{Neural Network Based Event Detection}
The state-of-the-art neural network models commonly transform event detection into a word-wise classification task.

Formally, let $D = \{(x_i,y_i)|i=1,2,...,n\}$ denote $n$ training instances, $P(y|x;\theta)$ is the neural network model parameterized by $\theta$, which takes representation (feature) $x$ as input and outputs the probability that $x$ is a trigger of event sub-type $y$ (or \textit{NIL}). Training procedure of such models commonly involves in minimizing following cross-entropy loss:

\begin{small}
\begin{equation}
\label{eq:ce}
\mathcal{L}_{CE}(\theta) = - \sum_{(x_i,y_i) \in D} \log P( y_i |x_i;\theta)
\end{equation}
\end{small}%
which corresponds to maximize the log-likelihood of the model making the correct prediction on all training instances and does not take the confusion between different type pairs into consideration.

\subsection{Cost-sensitive Regularization}
As discussed above, the key to improve event detection performance is to solve the label confusion problem, i.e., to guide the training procedure to concentrate on distinguishing between more confusing type pairs such as trigger/NIL pairs and sibling sub-event pairs. To this end, we propose cost-sensitive regularization, which reshapes the training loss with a cost-weighted term of the log-likelihood of making wrong prediction. Formally, the proposed regularizer is defined as:

\vspace{-0.3cm}
\begin{small}
\begin{equation}
\label{eq:ce}
\mathcal{L}_{CS}(\theta) = \sum_{(x_i,y_i) \in D} \sum_{y_j \neq y_i} C(y_i,y_j;x_i) \log P( y_j |x_i;\theta)
\end{equation}
\end{small}%
where $C(y_i,y_j;x)$ is a positive cost of mislabeling an instance $x$ with golden label $y_i$ into label $y_j$. A higher $C(y_i,y_j;x)$ is assigned if $y_i$ and $y_j$ is a more confusing type pair (i.e., more easily mislabeled by the current model). Therefore, the cost-sensitive regularizer will make the training procedure pay more attention to distinguish between confusing type pairs because they have larger impact on the training loss. Finally, the entire optimization objective can be written as:

\vspace{-0.1cm}
\begin{small}
\begin{equation}
\label{eq:ce}
\mathcal{L}(\theta) = \mathcal{L}_{CE}(\theta) + \lambda \mathcal{L}_{CS}(\theta)
\end{equation}
\end{small}%
where $\lambda$ is a hyper-parameter that controls the relative impact of our cost-sensitive regularizer.
\vspace{-0.2cm}
\section{Cost Estimation}
Obviously it is critical for the proposed cost-sensitive regularization to have an accurate estimation of the cost $C(y_i,y_j;x)$.
In this section, we propose two approaches for this issue based on population-level or instance-level statistics.

\subsection{Population-level Estimator}
A straightforward approach for measuring such costs is to use the relative mislabeling risk on the dataset. Therefore our population-level cost estimator is defined as:

\begin{small}
\begin{equation}
\label{eq:ce}
C_{POP}(y_i,y_j;x_i) = \frac{\#(y_i,y_j)}{\sum_j \#(y_i,y_j)}
\end{equation}
\end{small}%
where $\#(y_i,y_j)$ is the number of instances with golden label $y_i$ but being classified into class $y_j$ in the corpus. These statistics can be computed either on the training set or on the development set. This paper uses statistics on development set due to its compact size. And the estimators are updated every epoch during the training procedure.

\subsection{Instance-level Estimator}
The population-level estimators requires large computation cost to predict on the entire dataset when updating the estimators. To handle this issue, we propose another estimation method based directly on instance-level statistics. Inspire by~\citet{lin2017focal}, the probability $P(y_j|x_i;\theta)$ of classifying instance $x_i$ into the wrong class $y_j$ can be directly regarded as the mislabeling risk of that instance. Therefore our instance-level estimator is:

\begin{small}
\begin{equation}
\label{eq:ce}
C_{INS}(y_i,y_j;x_i) = P(y_j|x_i;\theta)
\end{equation}
\end{small}%
Then cost-sensitive regularizer for each training instance can be written as:

\begin{small}
\begin{equation}
\label{eq:ce}
\mathcal{L}_{INS}(x_i;\theta) = \sum_{y_j \neq y_i} P(y_j|x_i;\theta) \log P( y_j |x_i;\theta)
\end{equation}
\end{small}%
Note that if the probability of making correct prediction (i.e., $P(y_i|x_i;\theta)$) is fixed, $\mathcal{L}_{INS}(x_i;\theta)$ achieves its minimum when the probabilities of mislabeling $x_i$ into all incorrect classes are equal. This is equivalent to maximize the margin between the probability of golden label and that of any other class. In this circumstance, the loss $\mathcal{L}(\theta)$ can be regarded as a combination of maximizing both the likelihood of correct prediction and the margin between correct and incorrect classes.

\section{Experiments}
\subsection{Experimental Settings}
We conducted experiments on both English and Chinese on TAC-KBP 2017 Event Nugget Detection Evaluation datasets (LDC2017E55). For English, previously released RichERE corpus, including LDC2015E29, LDC2015E68, LDC2016E31 and the English part of LDC2017E02, were used for training. For Chinese, LDC2015E105, LDC2015E112, LDC2015E78 and the Chinese part of LDC2017E02 were used. For both English and Chinese, we sampled 20 documents from LDC2017E02 as the development set. Finally, there were 866/20/167 documents and 506/20/167 documents in English and Chinese train/development/test set respectively.

We conducted experiments on two state-of-the-art neural network event detection models to verify the portability of our method. One is DMCNN model proposed by~\citet{chen2015event}. Another is a LSTM model by~\citet{yang-mitchell:2017:Long}. Due to page limitation, please refer to original papers for details.

\subsection{Baselines\footnote{Our source code and hyper-parameter configures are openly available at \url{github.com/sanmusunrise/CSR}.}}
Following baselines were compared:

1) \textbf{Cross-entropy Loss (CE)}, the vanilla loss.

2) \textbf{Focal Loss (Focal)}~\cite{lin2017focal}, which is an instance-level method that rescales the loss with a factor proportional to the mislabeling probability to enhance the learning on hard instances.

3) \textbf{Hinge Loss (Hinge)}, which tries to separate the correct and incorrect predictions with a margin larger than a constant and is widely used in many machine learning tasks.

4) \textbf{Under-sampling (Sampling)}, a representative cost-sensitive learning approaches which samples instances balance the model learning and is widely used in event detection to deal with imbalance~\cite{chen2015event}.

We also compared our methods with the top systems in TAC-KBP 2017 Evaluation. We evaluated all systems with micro-averaged Precision(P), Recall(R) and F1 using the official  toolkit\footnote{\url{github.com/hunterhector/EvmEval}}.

\begin{table}[!t]
\setlength{\abovecaptionskip}{0.cm}
\setlength{\belowcaptionskip}{-0.5cm}
\begin{center}
\resizebox{0.49\textwidth}{!}{
\begin{tabular}{|l|c|c|c|c|c|c|}
  \hline
  \textbf{Model}          & \multicolumn{3}{c|}{\bf English} & \multicolumn{3}{c|}{\bf Chinese}   \\\cline{2-7}
                          & P             & R         & F1                & P         & R         & F1     \\ \hline
  \multicolumn{7}{|c|}{\bf LSTM}    \\ \hline
  CE                      & 73.46         & 34.23     & 46.70      & 70.35        & 35.43     & 47.13      \\ 
  Focal                   & 69.20         & 38.71     & 49.64      & 68.10        & 35.76     & 46.90       \\ 
  Hinge                   & 62.51         & 44.36     & 51.89      & 58.34        & 43.40     & 49.77        \\ 
  Sampling                & 58.57         & 48.26     & 52.92      & 57.61        & 44.54     & 50.24      \\ \hline
  CR-POP                  & 62.35         & 46.98     & 53.58      & 53.18        & 49.55     & 51.30       \\ 
  CR-INS                  & 58.64         & 49.55     & \textbf{53.71}      & 49.19        & 55.83     & \textbf{52.30}      \\ \hline
  \multicolumn{7}{|c|}{\bf DMCNN}    \\ \hline
  CE                      & 75.15         & 34.16     & 47.00      & 73.50        & 35.81     & 48.16     \\ 
  Focal                   & 70.68         & 37.63     & 49.11      & 69.04        & 38.87     & 49.74      \\ 
  Hinge                   & 67.49         & 42.67     & 52.28      & 60.27        & 45.50     & 51.85        \\ 
  Sampling                & 64.05         & 45.08     & 52.91      & 54.85        & 50.35     & 52.50      \\ \hline
  CR-POP                  & 64.82         & 45.73     & 53.63          & 55.89        & 50.81     & 53.23       \\ 
  CR-INS                  & 64.74         & 46.14     & \textbf{53.88}      & 54.91        & 51.93     & \textbf{53.38}      \\ \hline

\end{tabular}
}
\end{center}
\caption{Overall results. \emph{CR-POP} and \emph{CR-INS} are our method with population-level and instance-level estimators. All F1 improvements made by \emph{CR-POP} and \emph{CR-INS} are statistically significant with  $p < 0.05$. }
\label{tab:overall_rst}
\end{table}

\subsection{Overall Results}

Table~\ref{tab:overall_rst} shows the overall performance on TAC-KBP 2017 datasets. We can see that:

1) \textbf{Cost-sensitive regularization can significantly improve the event detection performance by taking mislabeling costs into consideration.} The proposed CR-INS and the CR-POP steadily outperform corresponding baselines. Besides, compared with population-level estimators, instance-level cost estimators are more effective. This may because instance-level estimators can be updated every batch while population-level estimators are updated every epoch, which leads to a more accurate estimation.

2) \textbf{Cost-sensitive regularization is robust to different languages and models.} We can see that cost-sensitive regularization achieves significant improvements on both English and Chinese datasets with both CNN and RNN models. This indicates that our method is robust and can be applied to different models and datasets.

3) \textbf{Data imbalance is not the only reason behind label confusion.} Even Focal and Sampling baselines deals with the data imbalance problem, they still cannot achieve comparable performance with CR-POP and CR-INS. This means that there are still other reasons which are not fully resolved by conventional methods for data imbalance.

\subsection{Comparing with State-of-the-art Systems}
\begin{figure}[!t]
\setlength{\belowcaptionskip}{-0.5cm}
  \centering
  \includegraphics[width=0.45\textwidth]{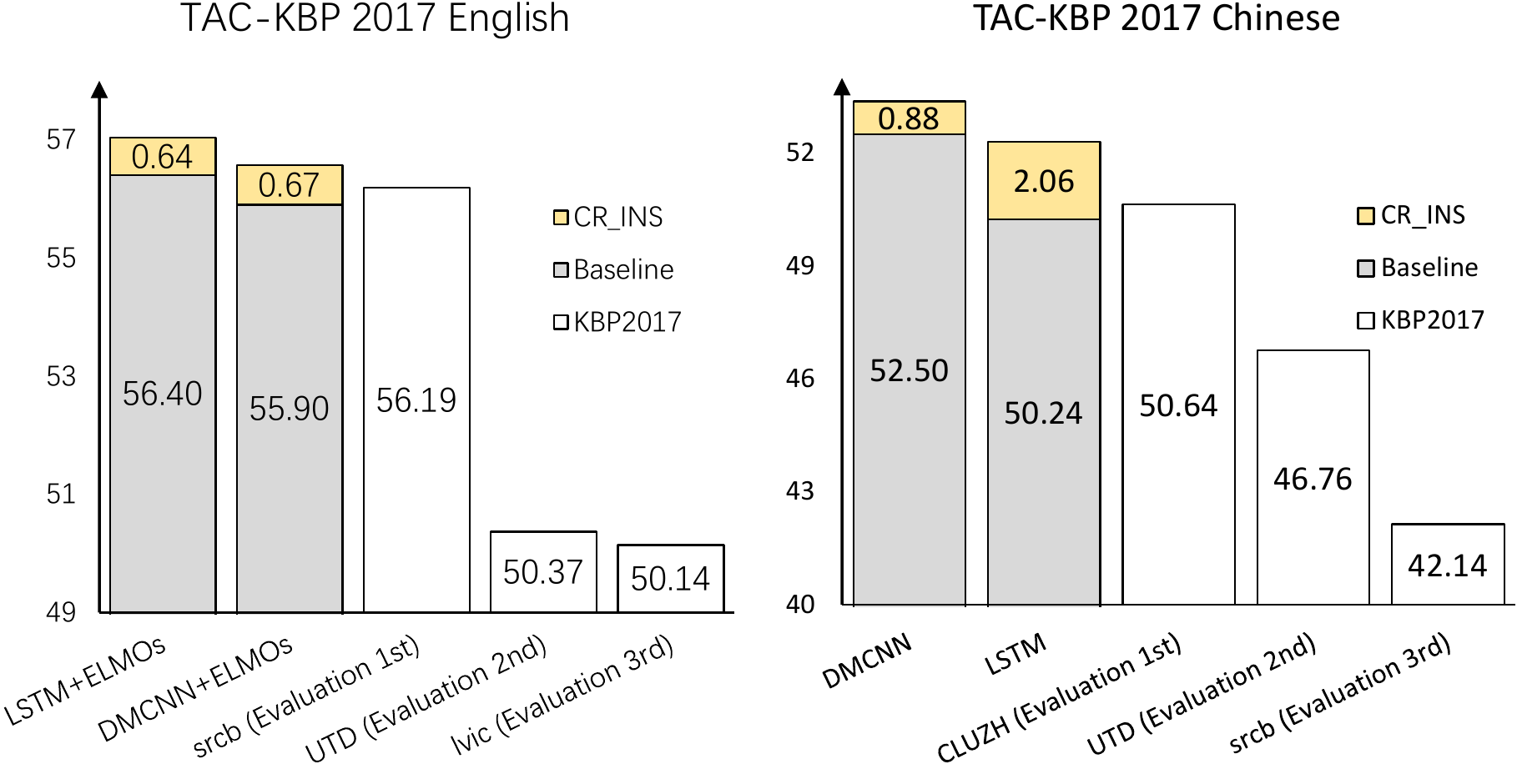}\\
  \caption{Comparison with the top systems in TAC-KBP 2017. CR is our CR-INS method. The \emph{srcb} system in English used additional CRF based models to deal with multi-word triggers in English, which is not considered in our model and leads to a significant higher recall than other competitors.}\label{fig:sota}
\end{figure}

Figure~\ref{fig:sota} compares our models with the top systems in TAC-KBP 2017 Evaluation. To achieve a strong baseline\footnote{Top systems in the evaluation are commonly ensembling models with additional resources, while reported in-house results are of single model.}, we also incorporate ELMOs~\cite{peters2018deep} to English system for better representations. We can see that CR-INS can further gain significant improvements over all strong baselines which have already achieved comparable performance with top systems. In both English and Chinese, CR-INS achieves the new SOTA  performance, which demonstrates its effectiveness.

\subsection{Error Analysis}

To clearly show where the improvement of our method comes from, we compared the mislabeling made by Sampling and our CR-INS method. Table~\ref{tab:err_ana} shows the results. We can first see that trigger/NIL mislabeling and sibling sub-types mislabeling make up most of errors of CE baseline. This further verifies our motivation. Besides, cost-sensitive regularization significantly reduces these two kinds of errors without introducing more other types of mislabeling, which clearly demonstrates the effectiveness of our method.

\begin{table}[!t]
\setlength{\abovecaptionskip}{0.cm}
\setlength{\belowcaptionskip}{-0.5cm}
\begin{center}
\resizebox{0.45\textwidth}{!}{
\begin{tabular}{l|r|r|r}
  \hline
  Error Rate (\%) & \multicolumn{1}{c|}{\textbf{SP}} & \multicolumn{1}{c|}{\textbf{CR}} & \multicolumn{1}{c}{\textbf{$\Delta$}} \\ \hline
  Total Error     			    &  42.97      &  38.84   &  -9.6\%\\ \hline
  - Trigger/NIL                 &  33.39	  &  31.15  &   -6.7\%  \\
  - Sibling Sub-types           &  8.15       &  6.25   &   -23.3\%  \\
  - Other                       &  1.43      &  1.44   &   +0.6\% \\ \hline

\end{tabular}
}
\end{center}
\caption{Error rates (CNN) on trigger words on the Chinese test set with Sampling(SP) and CR-INS(CR).}
\label{tab:err_ana}
\end{table}

\section{Related Work}

\noindent \textbf{Neural Network based Event Detection.} Recently, neural network based methods have achieved promising progress in event detection, especially with CNNs~\cite{chen2015event,nguyen2015event} and Bi-LSTMs~\cite{zeng2016convolution,yang-mitchell:2017:Long} based models as automatic feature extractors. Improvements have been made by incorporating arguments knowledge~\citep{Nguyen2016,liu-EtAl:2017:ACL,NguyenAndGrishman:2018:AAAI2018,hong-Etal:2018:ACL2018} or capturing larger scale of contexts with more complicated architectures~\citep{feng2016,nguyen2016a,ghaeini2016,lin2018adaptive,lin2018nugget,LiuEtal:2018:AAAI2018,Liu-and-Luo-and-Huang:2018:EMNLP2018,sha-Etal:2018:AAAI2018,Chen-Etal:2018:EMNLP2018}.

\noindent \textbf{Cost-sensitive Learning.}  Cost-sensitive learning has long been studied in machine learning~\cite{elkan2001foundations,zhou2011cost,ling2011cost}. It can be applied both at algorithm-level~\cite{anand1993improved,Domingos1999MetaCostAG,sun2007cost,krawczyk2014cost,kusner2014feature} or data-level~\cite{ting2002instance,zadrozny2003cost,mirza2013weighted}, which has achieved great success especially in learning with imbalanced data.

\section{Conclusions}
In this paper, we propose cost-sensitive regularization for neural event detection, which introduces a cost-weighted term of mislabeling likelihood to enhance the training procedure to concentrate more on confusing type pairs.  Experiments show that our methods significantly improve the performance of neural network event detection models.

\section*{Acknowledgments}

We sincerely thank the reviewers for their insightful comments and valuable suggestions. Moreover, this work is supported by the National Natural Science Foundation of China under Grants no. 61433015, 61572477 and 61772505; the Projects of the Chinese Language Committee under Grants no. WT135-24; and the Young Elite Scientists Sponsorship Program no. YESS20160177.

\bibliography{CSR}
\bibliographystyle{acl_natbib}

\end{document}